\documentclass[runningheads]{llncs}
\usepackage{graphicx}
\usepackage{cite}
\usepackage{amsmath,amssymb,amsfonts}
\usepackage{algorithmic}
\usepackage{textcomp}
\usepackage{xcolor}
\usepackage{subfig}
\usepackage{tabularray}
\usepackage{array}
\usepackage{colortbl}
\usepackage{fancyhdr}
\usepackage{multirow}
\usepackage{booktabs}
\begin{document}
\title{Hierarchical Skip Decoding for Efficient Autoregressive Text Generation}
%
%

\author{
Anonymous submission
\institute{}
}
\author{
Yunqi Zhu\inst{1,2}\orcidID{0000-0002-8147-3138} \and
Xuebing Yang\inst{2}\orcidID{0000-0001-8343-125X} \and
Yuanyuan Wu\inst{1}\orcidID{0000-0003-3409-2970} \and
Wensheng Zhang\inst{1,2,3}\orcidID{0000-0003-0752-941X}
}
\authorrunning{Yunqi et al.}
%
\institute{
School of Information and Communication Engineering, Hainan University, Haikou, China
\\
\email{ \{zhuyunqi96, wyuanyuan82\} @163.com}
\\
\email{zhangwenshengia@hotmail.com}
\and
Institute of Automation, Chinese Academy of Sciences, Beijing, China
\email{yangxuebing2013@ia.ac.cn}
\and
Guangzhou University, Guangzhou, China
}

\maketitle              

\begin{abstract}

Autoregressive decoding strategy is a commonly used method for text generation tasks with pre-trained language models,
while early-exiting is an effective approach to speedup the inference stage.
In this work, we propose a novel decoding strategy named Hierarchical Skip Decoding (HSD) for efficient autoregressive text generation. 
Different from existing methods that require additional trainable components, 
HSD is a plug-and-play method applicable to autoregressive text generation models,
it adaptively skips decoding layers in a hierarchical manner based on the current sequence length, 
thereby reducing computational workload and allocating computation resources. 
Comprehensive experiments on five text generation datasets with pre-trained language models demonstrate HSD's advantages in balancing efficiency and text quality. 
With almost half of the layers skipped, HSD can sustain 90\% of the text quality compared to vanilla autoregressive decoding, 
outperforming the competitive approaches.
\end{abstract}

\section{Introduction}

Large pre-trained language models, 
having dominated the field of natural language processing, 
have found extensive applications with prominent performances,
such as question answering, commonsense reasoning, summarization and machine translation \cite{devlin-etal-2019-bert, Radford2018ImprovingLU, Radford2019LanguageMA, raffel-2020-t5, chowdhery2022palm, Abdin2023phi2}.
Autoregressive decoding strategy for the language model has gained significant attention due to its effectiveness in generating coherent contexts. 
However, its efficiency, 
especially in terms of computational resources, 
has been a major challenge for researchers. 
To reduce the computational workload during inference, 
numerous approaches have been proposed, 
including early-exiting \cite{branchynet-Teerapittayanon, Adaptive-NN-effi-Bolukbasi, DeeBERT-Ji}, 
knowledge distillation \cite{hintonDistill2015, DynaBERT-Hou, pan-etal-2021-meta}, 
model pruning \cite{Srinivas2015DatafreePP, deepcomp2016han, li2017pruning, xia-etal-2022-structured}, 
sparse computation \cite{Kitaev2020Reformer, beltagy2020longformer, wang2020linformer}.

Early-exiting enables a model to terminate the computation early when a certain confidence threshold is met, 
thus bypassing the remaining layers. 
This concept has been explored in computer vision domains, 
where models like BranchyNet \cite{branchynet-Teerapittayanon} and SkipNet \cite{SkipNet-Wang} dynamically determine a possible exit point through external trained classifiers for the inference stage. 
Recently, 
early-exiting has been extended to the natural language processing field \cite{AdaptiveDepth-Elbayad, DeeBERT-Ji, BERxiT-xin}, 
with transformer-based models being fine-tuned with additional networks to facilitate the early-exiting.
In this study, 
we propose Hierarchical Skip Decoding (HSD), 
a novel approach that combines hierarchical layer skipping with scheduled computational budget. 
HSD will jump over the decoding layers in a hierarchical manner based on the current sequence length, 
thereby adaptively allocating computation resources. 
Unlike competitive methods with additional trainable components, 
HSD is a plug-and-play strategy that can be easily applied to any pre-trained language model,
and is compatible with the existing confidence score based methods.

\begin{figure}[t]
  \centering
  \subfloat[GPT-2]{
    \includegraphics[width=0.45 \linewidth]{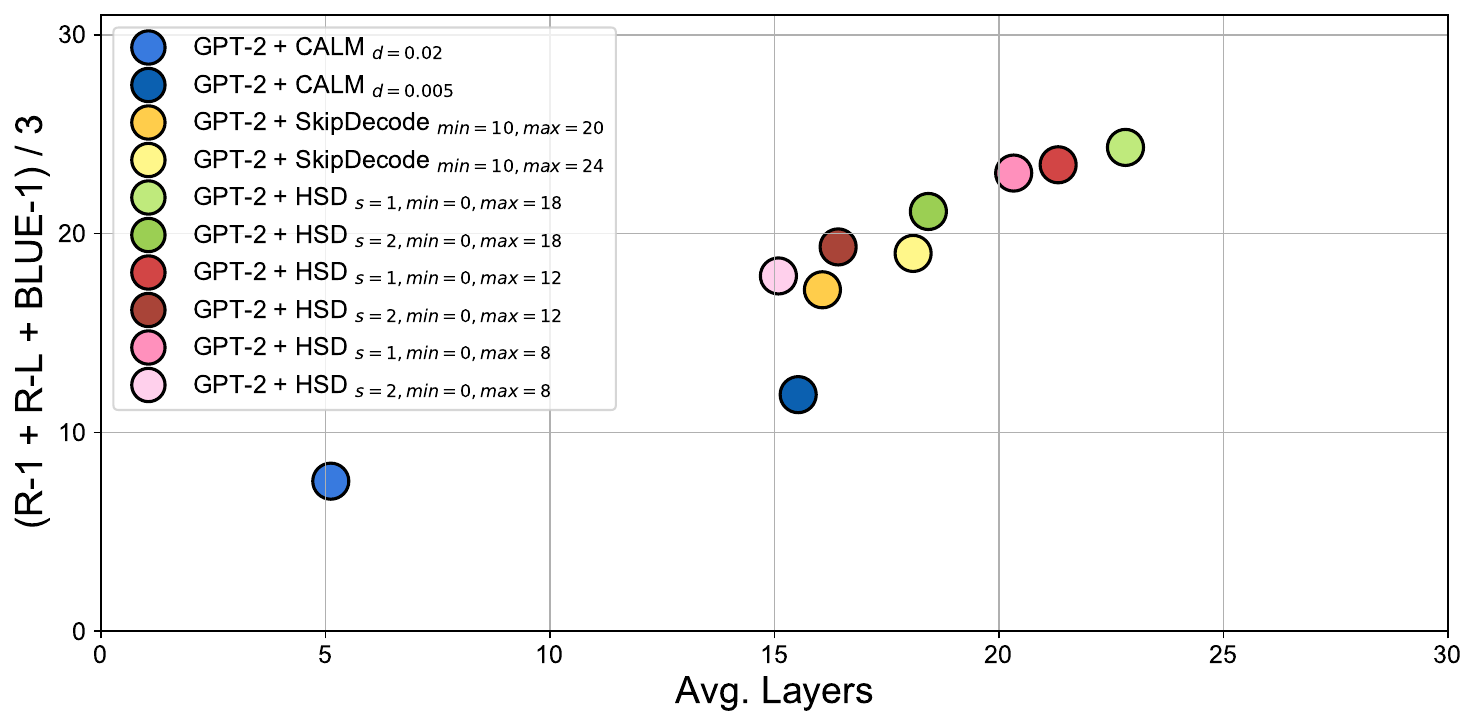}
  }
  \subfloat[Phi-2]{
    \includegraphics[width=0.45 \linewidth]{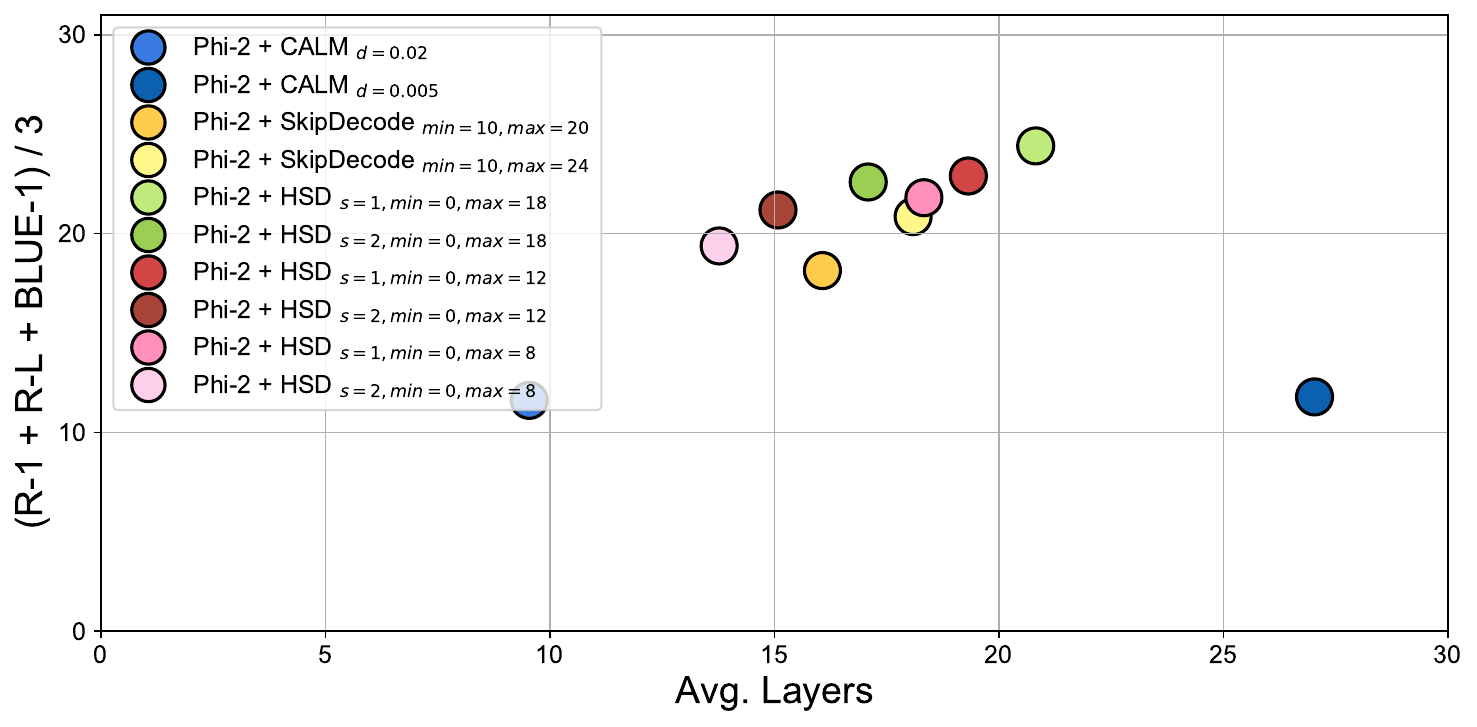}
    
  }
  \caption{
    Comparison of different methods in terms of evaluation results and number of decoding layers on CNN/DM.
  }
  \label{fig:fig1}
\end{figure}

To evaluate the effectiveness of HSD, we use two pre-trained transformer-based language models, GPT-2 \cite{Radford2019LanguageMA} and Phi-2 \cite{Abdin2023phi2}, and implement the experiments on five text generation datasets, covering news summarization, expansion of commonsense concept sets, and medical consultation domains. 
The results demonstrate HSD significantly outperforms existing methods, 
achieving a better balance between efficiency and text generation quality.
Fig~\ref{fig:fig1} visualize the performance comparison for the proposed method and the competitive methods.
Notably, with adaptively skipping almost half of the transformer-layers, 
HSD manages to preserve 90\% of the text quality compared to the original autoregressive decoding strategy, 
showcasing its advantages in maintaining text quality while reducing computational workload.
Overall, our study makes the following contributions: 
\textit{(i)} We introduce HSD, a hierarchical layer skipping and decoding strategy for efficient text generation. 
offering a flexible and plug-and-play solution, 
making it easy to deploy on pre-trained language models.
\textit{(ii)} We conduct comprehensive experiments on five datasets with two pre-trained language models, 
demonstrating HSD's superiority and robustness over existing methods.

\section{Related Work}

Autoregressive decoding is a commonly used text generation method for language models. 
In recent years, 
researchers have employed various techniques 
such as knowledge distillation \cite{hintonDistill2015, DynaBERT-Hou, pan-etal-2021-meta}, 
pruning \cite{Srinivas2015DatafreePP, deepcomp2016han, li2017pruning, xia-etal-2022-structured}, 
sparse computation \cite{Kitaev2020Reformer, beltagy2020longformer, wang2020linformer}, 
accelerating computation at the IO level \cite{dao2022flashattention, dao2023flashattention2}, 
storaging and retrieving of knowledge representations to skip computation \cite{lan2023copy, gim2023prompt, xiao2023efficient}, 
etc.
Additionally, 
efforts have been made to leverage RNN-like model design to reduce the memory footprint during inference \cite{peng2023rwkv, sun2023retentive}.

In this work, 
we concentrate on enhancing the inference speed by conditional layer skipping, 
while ensuring the generation quality by incorporating hierarchical skipping.
Early-exiting is an efficient approach for reducing computation based on whether the current hidden states meet a confident criteria for the incoming input.
The corresponding confidence score can be from vocabulary head projections, 
trained classifier's prediction, 
and cosine similarity to the previous layer's of the intermediate hiddent states etc.

For computer vision, BranchyNet \cite{branchynet-Teerapittayanon} is
a framework that introduces additional branches of convolutional blocks and classifiers at layers of different depths to determine the exit point.
Based on convolutional and residual neural networks, 
BranchyNet is trained from scratch and validated on computer vision datasets.
Further, Bolukbasi et al. \cite{Adaptive-NN-effi-Bolukbasi} transfers the idea of early-exiting into complex systems and environments with resource-rich cloud and resource-constrained edge devices.
Furthermore, trained with additional gating networks,
SkipNet \cite{SkipNet-Wang} focuses on learning how many convolutional blocks should be skipped rather than where to exit earlier.
Moreover, SDN \cite{SDN-Kaya}, an additional-classifier-based framework, includes feature reduction mechanism for supporting larger networks,
and is validated with pre-trained vision models.

For natural language processing,
Elbayad et al. \cite{AdaptiveDepth-Elbayad} verified the potential of early-exiting in autoregressive text generation with transformer-based model,
the research also showed that training all the early-exiting classifier at the end of each layer equally and simultaneously can be better than simulating autoregressive decoding strategy (i.e., merely training exited classifiers for each sampled token).
Further, DeeBERT \cite{DeeBERT-Ji} applied vocabulary head projections for every transformer-layer to minimize the distribution between the projected representations and the ground-truth token label with cross-entropy loss,
At the inference stage, a hyper-parameter threshold is included to decide the exit point if entropy of vocabulary head projections reaches the threshold.
With the rise of pre-trained language models,
FastBERT \cite{FastBERT-Liu} based on the pre-trained language model BERT, 
leverages the concept of self-distillation by incorporating classifiers into each layer of BERT
and reducing the predicted distribution of student classifiers and teacher classifiers during the fine-tuning stage.
Further, BERxiT \cite{BERxiT-xin} adopts fine-tuning strategies of \textit{(i)} equally optimized the early-exiting classifiers and the final classifier and \textit{(ii)} first updates the backbone model and the final classifier, and then fine-tunes the early-exiting classifiers with the frozen backbone model in an alternating fashion during training.
Additionally, BERxiT extends the early-exiting method to regression tasks.
In order to alleviate the issue of misjudgement by the early-exiting modules in the early layers, 
PABEE \cite{PABEE-zhou} applies the idea of cumulative voting mechanism to the inference stage.
Further early-exiting researches focus on improving the confidence measurement or difficulty measurement \cite{confident-Schwartz, CascadeBERT-li, HASHEE-sun},
hidden states copying \cite{FasterDAT-Liu}, 
hyper-parameter calibration framework \cite{confalm-Schuster},
and arranging layer-skipping in the early and middle layers by the current length of the generated sequence \cite{skipdec-Corro}.

\section{Methodology}

For the autoregressive sequence-to-sequence model,
denote $x$ as the input, $y$ as the target, $T$ as the current sequence length,
the probability of next token is predicted based on the previous given prompt and generated tokens,
\begin{equation}
  P_{\theta}(y|x) = \prod \limits_{t=1}^{T} P_{\theta}(y_{t}|y_{<t}, x).
  \label{eq:1}
\end{equation}
For the typical greedy search strategy,
each token is chosen by the highest probability,
\begin{equation}
  y_{t} = \mathop{\arg\max} P_{\theta}(y|x).
  \label{eq:2}
\end{equation}
For autoregressive decoding with early-exiting,
denote $L$ as the number of layers of the neural network,
$D$ as the vocabulary size, $d_{h} \in \mathbb{R}^{T \times d_{h}}$ as the hidden states size of the model,
$h_{t}^{i}$ as the intermediate hidden states of the current token $t$ in layer $i$,
$W_{lm} \in \mathbb{R}^{d_{h} \times D}$ as the final vocabulary weights of the language model,
the next token $y_{t+1}$ is chosen based on a confidence threshold $\lambda_{t}^{i}$ and a confidence score $C_{t}^{i}$,
\begin{equation}
  y_{t+1} = \mathop{\arg\max} (\mathrm{Softmax}(W_{lm} h_{t}^{i})), \mathrm{if\;} C_{t}^{i} \geq \lambda_{t}^{i}.
  \label{eq:3}
\end{equation}
where $C_{t}^{i}$ can be obtained by trained classifiers \cite{AdaptiveDepth-Elbayad, FastBERT-Liu, BERxiT-xin},
or calculating the significance of a token's hidden states or the cosine similarity of a token in the previous layer to the current layer using the intermediate variables $h_{t}^{i-1}$, $h_{t}^{i}$ or $\mathrm{Softmax}(W_{lm} h_{t}^{i})$ \cite{DeeBERT-Ji, confalm-Schuster}.

CALM \cite{confalm-Schuster} arranges a descending threshold $\lambda_{t}^{i}$ as the hidden states reach deeper layers.
However, once the misjudgement occurs in earlier tokens $t$ of the sequence,
the noise and perturbation could cause significant \textit{exposure bias} \cite{ranzato2016seq},
and consequentially devastating semantic and contextual errors.

SkipDecode \cite{skipdec-Corro} does not involves confidence score and early-exiting threshold the decoding,
but schedules budgets of computational resources for the generated sequences.
SkipDecode controls hyper-parameters of a minimum exit layer $E_{min}$ and a maximum exit layer $E_{max}$.
Denotes the maximum sequence length $T_{max}$, 
a descending number of scheduled forward layers will be applied for the current token:
\begin{equation}
  \{i \;|\; 1 \leq i \leq L, \; i \geq L - \frac{T \cdot E_{min} + (T_{max} - T) \cdot E_{max}}{T_{max}}\}.
  \label{eq:4}
\end{equation}

Hierarchical Skip Decoding (HSD) integrates the concepts of descending scheduled forward layers \cite{skipdec-Corro} and hierarchical layer skipping \cite{layerdrop, layer-dropping}.
For hierarchical layer-skipping, denotes the number of equally skipped layers as $s$, the scheduled foward layers can be:
\begin{equation}
  \{i \;|\; 1 \leq i \leq L, i \bmod (s + 1) = 0 \}.
  \label{eq:5}
\end{equation}
Since $\frac{s L}{s+1} $ layers will be skipped, 
a few layers can be redistributed by integrating with equation~\ref{eq:4},
so that the hyper-parameters both $E_{max}$ and $E_{min}$ can be smaller than the original SkipDecode\cite{skipdec-Corro}.

\section{Experiment}
In this section, 
we first introduce the statistic of the datasets.
Further we details the experiment settings and the compared methods.
Finally we elaborate and discuss the experimental result on the datasets, 

\subsection{Datasets}
We evaluate the proposed method on the datasets CNN/DM, XSum, E2E, CommonGen and HealthMagicCare.
\textit{(i)} CNN/DM \cite{nallapati-etal-2016-abstractive} is a News abstractive summarization dataset with a full articles as the document and a highlight as the reference summary.
\textit{(ii)} XSum \cite{narayan-etal-2018-dont} is a News abstractive summarization dataset with articles from BBC News and extremely abstractive one-sentence summaries as the output.
\textit{(iii)} E2E \cite{novikova-etal-2017-e2e} is a end-to-end text generation challenge for restaurant introduction with highly comdensed descriptive keywords as the input and human-written references as the output.
\textit{(iv)} CommonGen \cite{lin-etal-2020-commongen} is a commonsense reasoning task with a few scenario concepts-sets as the input and a relational scenario description as the output,
the datasets was collected from crowdsourcing annotations and visual captions. 
\textit{(v)} HealthMagicCare \cite{zeng-etal-2020-meddialog} is a patient-doctor dialogues dataset collected from an online medical consultation platform.
Statistics of the datasets are shown in Table~\ref{tab:dataset-stats},
\% novel N-gram is the proportion of unseen N-grams from the reference output over the input.

\begin{table}[htbp]
\centering
  \caption{Statistics of datasets.}
  \label{tab:dataset-stats}
    \begin{tabular}{lcrrrrcc}
      \toprule
      \multirow{2}{*}{Dataset} & \multirow{2}{*}{Train/Eval/Test} & \multicolumn{2}{c}{Avg Input} & \multicolumn{2}{c}{Avg Output} & \multicolumn{2}{c}{\% novel} \\
                         &                      & Sents     & Words         & Sents      & Words   & unigram  & bigram  \\
      \hline
      CNN/DM             & 287K / 13K / 11k     & 42.4      & 690.9         & 3.8        & 49.0    & 17       & 62      \\
      XSum               & 204K / 11K / 11k     & 19.0      & 430.3         & 1.0        & 23.2    & 38       & 84      \\
      E2E                & 42K / 4.6K / 4.6K    & 1.0       & 30.3          & 1.5        & 21.7    & 56       & 84      \\
      CommonGen          & 32K / 1K / 1.4K      & 1.0       & 3.2           & 1.0        & 10.8    & 78       & 99      \\
      HealthCareMagic    & 165K / 20K / 20K     & 4.9       & 89.9          & 6.6        & 90.7    & 80       & 98      \\
      \bottomrule
  \end{tabular}
\end{table}

\subsection{Experiment settings}
We implement the experiments the pre-trained language models GPT-2 \cite{Radford2019LanguageMA} and Phi-2 \cite{Abdin2023phi2}.
For GPT-2, we used a version with 774M parameters, 36 transformer layers, 
a hidden state of 1280, a attention heads of 20, 
and the vocabulary size of 50K.
Further, with 2.7B parameters and 32 transformer layers, 
Phi-2 is pre-trained with ``textbook quality'' web data,
it has a the hidden state size of 2560, a attention heads of 20, 
and the vocabulary size of 50K.
We full fine-tune GPT-2 and fine-tune Phi-2 with LoRA \cite{hu2022lora} over the five datasets respectively.
We fine-tune the models with AdamW optimizer and a learning rate of 5 $\times$ 10$^{-5}$ and 3 epochs.
We set the rank $r$ as 16 and factor $\alpha$ as 32 for LoRA.
We use beam search with a beam with of 4 in the inference stage.
All the experiments are conducted on a single NVIDIA A40 48GB GPU.

\subsection{Evaluation Metrics}
We evaluate the machine-generated texts with ROUGE and BLEU.
\textit{(i)} ROUGE \cite{lin-2004-rouge} is a recall-oriented evaluation metric that is well acknowledged in the realm of automatic summarization and is computationally efficient. 
It calculates the extent of intersection between the candidate and the reference at various levels of N-grams,
reflecting the ability of a system to retrieve key information.
The R-1, R-2, and R-L metrics represent the degree of concurrence between the unigram, bigram, and longest common subsequence of the candidate and the reference, respectively.
\textit{(ii)} BLEU \cite{papineni-2002-bleu} is a widely used evaluation metric for machine translation, question answering and automatic summarization.
It computes the precision of the overlapping N-grams between the candidate text and the reference text.
\textit{(iii)} Leveraging the pre-trained language model's ability to extract contextual semantics as hidden representations,
BERTScore \cite{Zhang2020BERTScore} computes the cosine similarity between the reference text and the generated text.

\subsection{Compared methods}

We compared the propose method with the conventional full fine-tuning, 
CALM \cite{confalm-Schuster} and SkipDecode \cite{skipdec-Corro} respectively on the pre-trained text generation model GPT-2 and Phi-2.

For CALM, 
the threshold $\lambda_{t}^{i}$ is a decreasing series as the hidden representation passes through deeper layers $i$ and is controlled by a hyperparameter $d$, 
where we set $d$ to 0.02 and 0.005 in this study and denote the the corresponding method as CALM $_{d=0.02}$ and CALM $_{d=0.005}$ respectively.
Further,
refered as ``softmax response" in CALM's study \cite{confalm-Schuster},
we measure the confidence score by the gap between the two largest values of the vocabulary projection of the current layer's hidden representations.

For SkipDecode, 
it gets rid of calculating the confidence scores with intermediate variables and hidden representations and saves computing resources,
where we set the minimum and maximum exit layer ($E_{min}$, $E_{max}$) to as (10, 20) and (10, 24) in this study and denote the corresponding method as SkipDecode $_{min=10, max=20}$ and SkipDecode $_{min=10, max=24}$ respectively for the comparison.

\subsection{Experimental result}

As shown in the following Table \ref{tab:CNNDM}, \ref{tab:XSum}, \ref{tab:E2E}, \ref{tab:CommonGen} and \ref{tab:HealthMagicCare}, 
the experimental results shows that CALM is sensitive to the hyperparameter $d$, 
the difference between $d$ = 0.02 and 0.05 is considerably large in terms of the average passing layers in the model and text quality,
which generally fails to achieve the balance between saving computational resources and perserving text generation quality.

Further, 
both SkipDecode and HSD adopt the scheduled decoding strategy, 
therefore the average number of passing layers is relatively stable on different datasets.
In terms of the trade-off between being energy-efficient and maintaining ROUGE, BLUE and BERTScore,
the proposed method yields more equitable advantages, 
exemplified by its superior performance with similar number of computing layers (e.g., HSD $_{s=1, min=0, max=8}$ vs. SkipDecode $_{min=10, max=24}$ and HSD $_{s=2, min=0, max=12}$ vs. SkipDecode $_{min=10, max=24}$).
Furthermore, 
comparing to the vanilla autoregressive decoding strategy,
although roughly 40-60\% of layers are discarded during the inference stage,
the proposed method achieves perserving about 70-90\% of ROUGE scores and BLUE-1 on the five datasets.
Hence, 
hyperparameters of $s \in \{ 1, 2 \}$, $min = 0$ and $max \in [ 12, 18 ]$ can be the sweet point for the proposed method.
It is also observed that the evaluation results of BERTScore are relatively volatile that may more objectively reflect the text generation performance,
but the proposed method accomplishes a superior performance over the compared methods.
Overall, 
the performance and feasibility of the proposed approach is validated across five datasets, 
covering news summarization, 
end-to-end text generation, 
contextual commonsense reasoning and generation,
and medical dialogue generation.

\definecolor{cgray}{RGB}{240,240,240}
\begin{table}[!htbp]
\centering
  \caption{CNN/DM}
  \label{tab:CNNDM}
  \scalebox{0.73}{
  \begin{tabular}{lrrrrrc}
    \toprule
      & Avg. Layer & R-1 & R-2 & R-L & BLEU-1 & BERTScore  \\
    \midrule
    GPT-2 & 36.00 & 29.95 & 8.03 & 27.98 & 21.79 & 56.26 \\
    \midrule
    + CALM $_{d=0.02}$ & 5.12 & 6.85 & 0.71 & 6.38 & 9.40 & 28.67 \\
    + CALM $_{d=0.005}$ & 15.53 & 12.10 & 0.95 & 11.10 & 12.49 & 33.36 \\
    + SkipDecode $_{min=10, max=20}$ & 16.07 & 19.09 & 2.91 & 17.70 & 14.73 & 34.90 \\
    + SkipDecode $_{min=10, max=24}$ & 18.09 & 21.17 & 4.12 & 19.70 & 16.14 & 38.11 \\
    \rowcolor{cgray} + HSD $_{s=1, min=0, max=18}$ & 22.82 & 27.25 & 6.92 & 25.38 & 20.38 & 51.28 \\
    \rowcolor{cgray} + HSD $_{s=2, min=0, max=18}$ & 18.43 & 23.45 & 5.22 & 21.56 & 18.33 & 42.45 \\
    \rowcolor{cgray} + HSD $_{s=1, min=0, max=12}$ & 21.32 & 26.19 & 6.36 & 24.32 & 19.88 & 49.95 \\
    \rowcolor{cgray} + HSD $_{s=2, min=0, max=12}$ & 16.42 & 21.15 & 4.06 & 19.43 & 17.41 & 39.07 \\
    \rowcolor{cgray} + HSD $_{s=1, min=0, max=8}$  & 20.33 & 25.66 & 6.10 & 23.76 & 19.72 & 49.23 \\
    \rowcolor{cgray} + HSD $_{s=2, min=0, max=8}$  & 15.09 & 19.23 & 3.18 & 17.66 & 16.72 & 36.55 \\
    \midrule
    Phi-2 & 32.00 & 31.63 & 8.82 & 29.39 & 22.71 & 57.93 \\
    \midrule
    + CALM $_{d=0.02}$ & 9.54 & 12.29 & 0.87 & 11.11 & 11.43 & 31.70 \\
    + CALM $_{d=0.005}$ & 27.03 & 12.77 & 0.93 & 11.87 & 10.71 & 33.14 \\  
    + SkipDecode $_{min=10, max=20}$ & 16.07 & 20.47 & 3.47 & 19.40 & 14.55 & 40.34 \\
    + SkipDecode $_{min=10, max=24}$ & 18.09 & 23.68 & 5.22 & 22.36 & 16.53 & 44.73 \\
    \rowcolor{cgray} + HSD $_{s=1, min=0, max=18}$ & 20.82 & 27.44 & 7.09 & 25.69 & 20.11 & 51.83 \\
    \rowcolor{cgray} + HSD $_{s=2, min=0, max=18}$ & 17.09 & 25.06 & 5.71 & 23.40 & 19.33 & 46.99 \\
    \rowcolor{cgray} + HSD $_{s=1, min=0, max=12}$ & 19.32 & 25.61 & 6.23 & 23.99 & 19.10 & 50.02 \\
    \rowcolor{cgray} + HSD $_{s=2, min=0, max=12}$ & 15.08 & 23.25 & 4.58 & 21.70 & 18.62 & 43.46 \\
    \rowcolor{cgray} + HSD $_{s=1, min=0, max=8}$  & 18.33 & 24.31 & 5.68 & 22.77 & 18.34 & 48.55 \\
    \rowcolor{cgray} + HSD $_{s=2, min=0, max=8}$  & 13.77 & 20.99 & 3.35 & 19.54 & 17.61 & 40.51 \\
    \bottomrule
  \end{tabular}
}
\end{table}

\begin{table}[!htbp]
\centering
  \caption{XSum}
  \label{tab:XSum}
  \scalebox{0.73}{
  \begin{tabular}{lrrrrrc}
    \toprule
      & Avg. Layer & R-1 & R-2 & R-L & BLEU-1 & BERTScore  \\
    \midrule
    GPT-2 & 36.00 & 18.05 & 6.04 & 15.48 & 10.08 & 55.31 \\
    \midrule
    + CALM $_{d=0.02}$ & 5.06 & 7.01 & 1.19 & 6.12 & 4.31 & 26.44 \\
    + CALM $_{d=0.005}$ & 14.93 & 10.25 & 1.30 & 8.53 & 6.74 & 30.29 \\
    + SkipDecode $_{min=10, max=20}$ & 16.07 & 13.03 & 2.97 & 11.18 & 7.46 & 34.39 \\ 
    + SkipDecode $_{min=10, max=24}$ & 18.09 & 14.34 & 4.00 & 12.33 & 8.20 & 36.73 \\
    \rowcolor{cgray} + HSD $_{s=1, min=0, max=18}$ & 22.82 & 17.48 & 5.89 & 15.03 & 9.96 & 52.05 \\
    \rowcolor{cgray} + HSD $_{s=2, min=0, max=18}$ & 18.43 & 16.32 & 5.02 & 13.67 & 9.93 & 42.87 \\
    \rowcolor{cgray} + HSD $_{s=1, min=0, max=12}$ & 21.32 & 16.81 & 5.35 & 14.40 & 9.65 & 50.93 \\
    \rowcolor{cgray} + HSD $_{s=2, min=0, max=12}$ & 16.42 & 14.70 & 3.77 & 12.27 & 9.19 & 38.91 \\
    \rowcolor{cgray} + HSD $_{s=1, min=0, max=8}$  & 20.33 & 16.36 & 4.93 & 13.97 & 9.40 & 50.18 \\
    \rowcolor{cgray} + HSD $_{s=2, min=0, max=8}$  & 15.09 & 13.32 & 2.91 & 11.14 & 8.62 & 35.85 \\
    \midrule
    Phi-2 & 32.00 & 18.53 & 6.11 & 15.88 & 10.3 & 56.09 \\
    \midrule
    + CALM $_{d=0.02}$ & 9.54 & 9.95 & 1.22 & 8.24 & 6.26 & 31.26 \\
    + CALM $_{d=0.005}$ & 27.16 & 10.6 & 1.46 & 9.15 & 5.63 & 31.43 \\
    + SkipDecode $_{min=10, max=20}$ & 16.07 & 14.43 & 3.40 & 12.49 & 8.33 & 39.37 \\
    + SkipDecode $_{min=10, max=24}$ & 18.09 & 17.07 & 5.03 & 14.67 & 9.51 & 44.01 \\
    \rowcolor{cgray} + HSD $_{s=1, min=0, max=18}$ & 20.82 & 17.76 & 5.76 & 15.18 & 9.76 & 51.82 \\
    \rowcolor{cgray} + HSD $_{s=2, min=0, max=18}$ & 17.09 & 17.49 & 5.21 & 14.81 & 10.38 & 47.89 \\
    \rowcolor{cgray} + HSD $_{s=1, min=0, max=12}$ & 19.32 & 16.75 & 4.93 & 14.27 & 9.19 & 50.33 \\
    \rowcolor{cgray} + HSD $_{s=2, min=0, max=12}$ & 15.08 & 16.08 & 3.93 & 13.44 & 9.91 & 44.24 \\
    \rowcolor{cgray} + HSD $_{s=1, min=0, max=8}$  & 18.33 & 15.67 & 4.22 & 13.33 & 8.76 & 49.66 \\
    \rowcolor{cgray} + HSD $_{s=2, min=0, max=8}$  & 13.77 & 15.03 & 2.96 & 12.25 & 9.80 & 41.06 \\
    \bottomrule
  \end{tabular}
  }
\end{table}

\begin{table}[!htbp]
\centering
  \caption{E2E}
  \label{tab:E2E}
  \scalebox{0.73}{
  \begin{tabular}{lrrrrrc}
    \toprule
      & Avg. Layer & R-1 & R-2 & R-L & BLEU-1 & BERTScore  \\
    \midrule
    GPT-2 & 36.00 & 56.27 & 30.24 & 48.51 & 41.34 & 70.68 \\
    \midrule
    + CALM $_{d=0.02}$ & 6.62 & 16.13 & 5.52 & 14.26 & 16.92 & 31.08 \\
    + CALM $_{d=0.005}$ & 13.08 & 18.70 & 5.30 & 16.10 & 18.77 & 36.56 \\
    + SkipDecode $_{min=10, max=20}$ & 16.20 & 38.43 & 18.09 & 32.96 & 29.37 & 50.99 \\
    + SkipDecode $_{min=10, max=24}$ & 18.28 & 41.55 & 20.81 & 35.14 & 31.35 & 53.18 \\
    \rowcolor{cgray} + HSD $_{s=1, min=0, max=18}$ & 22.94 & 56.38 & 31.99 & 49.53 & 42.15 & 69.06 \\
    \rowcolor{cgray} + HSD $_{s=2, min=0, max=18}$ & 18.57 & 52.44 & 29.73 & 46.87 & 39.35 & 64.36 \\
    \rowcolor{cgray} + HSD $_{s=1, min=0, max=12}$ & 21.38 & 56.38 & 31.98 & 49.45 & 42.45 & 68.84 \\
    \rowcolor{cgray} + HSD $_{s=2, min=0, max=12}$ & 16.51 & 50.58 & 27.98 & 45.16 & 38.29 & 64.23 \\
    \rowcolor{cgray} + HSD $_{s=1, min=0, max=8}$  & 20.36 & 55.89 & 31.56 & 48.92 & 42.04 & 69.06 \\ 
    \rowcolor{cgray} + HSD $_{s=2, min=0, max=8}$  & 15.16 & 50.13 & 27.37 & 44.77 & 38.01 & 64.41 \\ 
    \midrule
    Phi-2 & 32.00 & 58.91 & 32.97 & 50.47 & 42.62 & 69.69 \\
    \midrule
    + CALM $_{d=0.02}$ & 21.99 & 16.43 & 5.53 & 14.62 & 13.85 & 35.37 \\
    + CALM $_{d=0.005}$ & 27.86 & 19.41 & 6.65 & 17.17 & 15.05 & 37.06 \\
    + SkipDecode $_{min=10, max=20}$ & 16.20 & 40.39 & 18.05 & 35.56 & 31.93 & 53.78 \\
    + SkipDecode $_{min=10, max=24}$ & 18.28 & 47.41 & 23.57 & 41.89 & 35.85 & 60.12 \\
    \rowcolor{cgray} + HSD $_{s=1, min=0, max=18}$ & 20.94 & 56.54 & 30.07 & 49.45 & 41.32 & 70.16 \\
    \rowcolor{cgray} + HSD $_{s=2, min=0, max=18}$ & 17.24 & 52.68 & 26.11 & 46.39 & 40.16 & 64.45 \\
    \rowcolor{cgray} + HSD $_{s=1, min=0, max=12}$ & 19.38 & 54.22 & 28.57 & 47.61 & 39.98 & 68.41 \\
    \rowcolor{cgray} + HSD $_{s=2, min=0, max=12}$ & 15.18 & 49.67 & 23.10 & 43.17 & 38.77 & 61.89 \\
    \rowcolor{cgray} + HSD $_{s=1, min=0, max=8}$  & 18.36 & 52.88 & 26.11 & 46.25 & 38.53 & 67.19 \\
    \rowcolor{cgray} + HSD $_{s=2, min=0, max=8}$  & 13.83 & 46.80 & 20.31 & 40.53 & 37.08 & 59.45 \\
    \bottomrule
  \end{tabular}
  }
\end{table}

\begin{table}[!htbp]
\centering
  \caption{CommonGen}
  \label{tab:CommonGen}
  \scalebox{0.73}{
  \begin{tabular}{lrrrrrc}
    \toprule
      & Avg. Layer & R-1 & R-2 & R-L & BLEU-1 & BERTScore  \\
    \midrule
    GPT-2 & 36.00 & 38.69 & 13.55 & 32.48 & 24.73 & 60.53 \\
    \midrule
    + CALM $_{d=0.02}$ & 9.72 & 11.68 & 1.73 & 10.65 & 7.23 & 34.12 \\
    + CALM $_{d=0.005}$ & 16.77 & 13.97 & 2.16 & 12.34 & 8.58 & 38.48 \\
    + SkipDecode $_{min=10, max=20}$ & 16.34 & 27.64 & 6.39 & 21.84 & 15.64 & 45.58 \\  
    + SkipDecode $_{min=10, max=24}$ & 18.48 & 31.07 & 9.57 & 24.67 & 17.44 & 49.90 \\
    \rowcolor{cgray} + HSD $_{s=1, min=0, max=18}$ & 23.03 & 37.40 & 12.82 & 31.08 & 23.96 & 57.65 \\
    \rowcolor{cgray} + HSD $_{s=2, min=0, max=18}$ & 18.79 & 33.57 & 10.89 & 27.92 & 21.19 & 54.27 \\
    \rowcolor{cgray} + HSD $_{s=1, min=0, max=12}$ & 21.41 & 36.82 & 12.11 & 30.51 & 23.65 & 57.34 \\
    \rowcolor{cgray} + HSD $_{s=2, min=0, max=12}$ & 16.62 & 31.43 &  9.88 & 26.21 & 20.21 & 52.16 \\
    \rowcolor{cgray} + HSD $_{s=1, min=0, max=8}$  & 20.41 & 36.03 & 11.53 & 29.69 & 23.21 & 56.96 \\  
    \rowcolor{cgray} + HSD $_{s=2, min=0, max=8}$  & 15.20 & 30.76 &  9.73 & 25.68 & 19.83 & 51.74 \\   
    \midrule
    Phi-2 & 32.00 & 12.61 & 0.98 & 11.27 & 9.57 & 46.52 \\
    \midrule
    + CALM $_{d=0.02}$ & 25.55 & 8.65 & 0.35 & 7.89 & 5.56 & 36.01 \\
    + CALM $_{d=0.005}$ & 29.33 & 10.78 & 0.49 & 9.71 & 6.36 & 38.22 \\
    + SkipDecode $_{min=10, max=20}$ & 16.34 & 13.93 & 1.05 & 12.38 & 9.78 & 44.29 \\
    + SkipDecode $_{min=10, max=24}$ & 18.48 & 14.17 & 1.09 & 12.65 & 10.41 & 45.11 \\
    \rowcolor{cgray} + HSD $_{s=1, min=0, max=18}$ & 21.03 & 14.31 & 1.26 & 12.75 & 10.46 & 47.21 \\
    \rowcolor{cgray} + HSD $_{s=2, min=0, max=18}$ & 17.38 & 15.52 & 1.31 & 13.66 & 11.62 & 44.89 \\
    \rowcolor{cgray} + HSD $_{s=1, min=0, max=12}$ & 19.41 & 14.41 & 1.30 & 12.83 & 10.90 & 47.55 \\
    \rowcolor{cgray} + HSD $_{s=2, min=0, max=12}$ & 15.27 & 15.74 & 1.28 & 13.63 & 12.03 & 45.00 \\
    \rowcolor{cgray} + HSD $_{s=1, min=0, max=8}$  & 18.41 & 14.63 & 1.29 & 12.93 & 11.14 & 47.80 \\
    \rowcolor{cgray} + HSD $_{s=2, min=0, max=8}$  & 13.86 & 15.11 & 1.20 & 13.24 & 11.87 & 44.86 \\
    \bottomrule
  \end{tabular}
  }
\end{table}

\begin{table}[!htbp]
\centering
  \caption{HealthMagicCare}
  \label{tab:HealthMagicCare}
  \scalebox{0.73}{
  \begin{tabular}{lrrrrrc}
    \toprule
      & Avg. Layer & R-1 & R-2 & R-L & BLEU-1 & BERTScore  \\
    \midrule
    GPT-2 & 36.00 & 26.53 & 6.62 & 24.76 & 25.14 & 54.04 \\
    \midrule
    + CALM $_{d=0.02}$ & 4.94 & 7.48 & 0.64 & 6.80 & 8.86 & 28.84 \\
    + CALM $_{d=0.005}$ & 14.33 & 14.71 & 1.23 & 13.09 & 17.05 & 34.78 \\
    + SkipDecode $_{min=10, max=20}$ & 16.07 & 22.39 & 4.10 & 20.98 & 22.01 & 37.43 \\
    + SkipDecode $_{min=10, max=24}$ & 18.09 & 22.42 & 4.26 & 20.86 & 22.11 & 38.37 \\
    \rowcolor{cgray} + HSD $_{s=1, min=0, max=18}$ & 22.82 & 26.61 & 6.95 & 25.02 & 25.12 & 52.89 \\
    \rowcolor{cgray} + HSD $_{s=2, min=0, max=18}$ & 18.43 & 25.63 & 5.95 & 24.15 & 24.80 & 48.12 \\
    \rowcolor{cgray} + HSD $_{s=1, min=0, max=12}$ & 21.32 & 27.07 & 7.06 & 25.43 & 25.59 & 52.72 \\
    \rowcolor{cgray} + HSD $_{s=2, min=0, max=12}$ & 16.42 & 24.14 & 5.14 & 22.68 & 24.16 & 44.63 \\
    \rowcolor{cgray} + HSD $_{s=1, min=0, max=8}$  & 20.33 & 26.48 & 6.77 & 24.89 & 25.22 & 52.27 \\  
    \rowcolor{cgray} + HSD $_{s=2, min=0, max=8}$  & 15.09 & 22.69 & 4.52 & 21.37 & 23.24 & 42.05 \\
    \midrule
    Phi-2 & 32.00 & 27.59 & 6.39 & 25.86 & 24.05 & 54.70 \\
    \midrule
    + CALM $_{d=0.02}$ & 8.84 & 11.02 & 0.52 & 9.83 & 15.74 & 32.96 \\
    + CALM $_{d=0.005}$ & 26.00 & 13.05 & 0.59 & 11.76 & 15.12 & 34.18 \\
    + SkipDecode $_{min=10, max=20}$ & 16.07 & 22.24 & 3.60 & 21.00 & 23.18 & 43.17 \\
    + SkipDecode $_{min=10, max=24}$ & 18.09 & 23.42 & 4.36 & 22.05 & 23.01 & 44.98 \\
    \rowcolor{cgray} + HSD $_{s=1, min=0, max=18}$ & 20.82 & 24.70 & 4.77 & 23.13 & 20.47 & 48.55 \\
    \rowcolor{cgray} + HSD $_{s=2, min=0, max=18}$ & 17.09 & 24.42 & 4.31 & 22.84 & 23.14 & 46.02 \\
    \rowcolor{cgray} + HSD $_{s=1, min=0, max=12}$ & 19.32 & 24.37 & 4.54 & 22.92 & 21.27 & 47.48 \\
    \rowcolor{cgray} + HSD $_{s=2, min=0, max=12}$ & 15.08 & 22.75 & 3.40 & 21.36 & 23.13 & 42.48 \\
    \rowcolor{cgray} + HSD $_{s=1, min=0, max=8}$  & 18.33 & 23.74 & 4.07 & 22.26 & 19.24 & 45.61 \\
    \rowcolor{cgray} + HSD $_{s=2, min=0, max=8}$  & 13.77 & 19.60 & 2.14 & 18.48 & 20.92 & 38.99 \\
    \bottomrule
  \end{tabular}
  }
\end{table}

\section{Conclusion}

In this study,
we proposed a hierarchical layer skipping and decoding strategy called HSD for the language models in the inference stage. 
Rather than directly tailoring the original language model,
HSD skips the decoding layer in a scheduled and hierarchical manner, 
and strives for enhancing the overall text generation quality while utilizing a similar level of computational resources.
Evaluating on five text generation datasets and two pre-trained language models GPT-2 and Phi-2,
extensive experimental evidences show that the proposed method achieves compelling performance over the competitive methods in terms of efficacy and practicality.
By alleviating the efficiency challenge of autoregressive text generation, 
this work paves the way for the research of computationally efficient language models, 
thereby facilitating their broader adoption in practical applications. 

%
%
%
\bibliographystyle{splncs04}
\bibliography{prcv-2024}
%





\end{document}